\documentclass[letterpaper, 10 pt, conference,]{ieeeconf}% Comment this line out
\IEEEoverridecommandlockouts                              % This command is only
\usepackage{graphicx}
\usepackage{siunitx}
\usepackage[utf8]{inputenc}
\usepackage[T1]{fontenc}
\usepackage{amsfonts}
\newcommand{\fig}[1]{Fig.~\ref{#1}}

\usepackage{epsfig} % for postscript graphics files
\usepackage{amsmath} % assumes amsmath package installed
\usepackage{amssymb}  % assumes amsmath package installed
\usepackage{subcaption}
\usepackage{caption}
\usepackage{color}
\usepackage{verbatim}
\usepackage{wasysym}
\usepackage[table,xcdraw]{xcolor}
\usepackage{graphicx}
\usepackage{gensymb}
\usepackage{xcolor}
\usepackage{latexsym}
\usepackage{multirow}
\usepackage{amsmath}
\usepackage{booktabs,caption}
\usepackage[noadjust]{cite}
\usepackage[flushleft]{threeparttable}
\usepackage[table,xcdraw]{xcolor}

\usepackage[normalem]{ulem} % this prevent ulem to redefine \emph command 
\usepackage{wasysym}

\newcommand{\RNum}[1]{\uppercase\expandafter{\romannumeral #1\relax}}
\makeatletter
\newlength\tmp@\newlength\t@mp
\newcommand{\comp}[3]
  {\mathop{ \settowidth\tmp@{$\displaystyle\mathop{#1}^{#3}_{#2}$}
  \hbox to \tmp@{\hss \settowidth\t@mp{$\displaystyle #1$}\setlength\t@mp{.45\t@mp}
  $\displaystyle\mathop{#1}^{\hspace\t@mp #3}_{\hspace{-\t@mp}#2}$
  \hss} }}

\makeatother
\title{\LARGE \bf
Mechanisms and Computational Design of\\Multi-Modal End-Effector with Force Sensing using Gated Networks\\
}
\author{Yusuke Tanaka$^{1*}$, Alvin Zhu$^{2,3*}$, Richard Lin$^{3}$, Ankur Mehta$^{3}$ and Dennis Hong$^{1}$% <-this % stops a space
\thanks{$^{1}$Y. Tanaka and D. Hong are with the Department of Mechanical and Aerospace Engineering, UCLA, Los Angeles, CA, USA.}%
\thanks{$^{2}$A. Zhu is with the Department of Computer Science, UCLA, Los Angeles, CA, USA.}%
\thanks{$^{3}$R. Lin and A. Zhu are with the Electrical and Computer Engineering Department, UCLA, Los Angeles, CA, USA.}%
\thanks{Emails: {\tt\small yusuketanaka, alvister88, richardlin, dennishong@ucla.edu}}%
}
\begin{document}
 \twocolumn[{%
 \renewcommand\twocolumn[1][]{#1}%
     \maketitle
     \begin{center}
         \centering
         \includegraphics[width=0.9\textwidth]{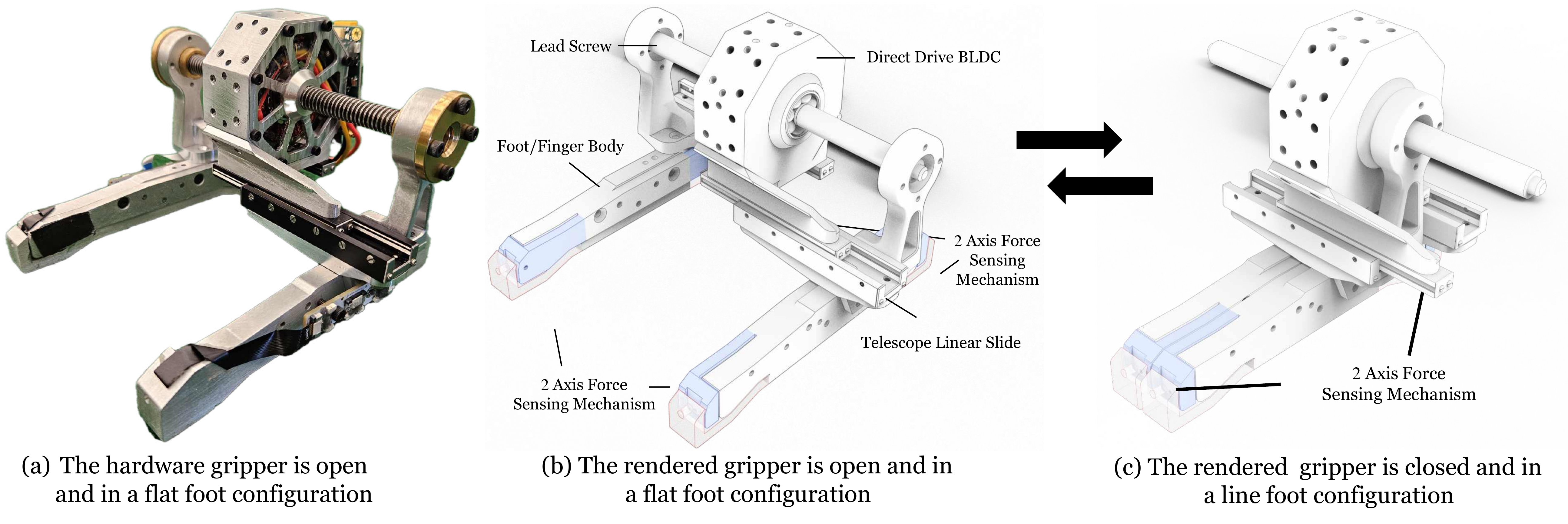} % 
     \captionof{figure}{MAGPIE Hardware and Design Rendering in an isometric view.}
     \label{fig:fig1}
     \end{center}%
     }]
         \footnotetext[1]{$^{1}$Y. Tanaka and D. Hong are with the Department of Mechanical and Aerospace Engineering, UCLA, Los Angeles, CA, USA. $^{2}$A. Zhu is with the Department of Computer Science, UCLA, Los Angeles, CA, USA. $^{3}$R. Lin, A. Mehta, and A. Zhu are with the Electrical and Computer Engineering Department, UCLA, Los Angeles, CA, USA. {\tt\small\{yusuketanaka, alvister88, richardlin, mehtank, dennishong\}@g.ucla.edu.} $^*$Y. Tanaka and A. Zhu assert joint first authorship.}%

\thispagestyle{empty}
\pagestyle{empty}

\begin{abstract}
In limbed robotics, end-effectors must serve dual functions, such as both feet for locomotion and grippers for grasping, which presents design challenges. This paper introduces a multi-modal end-effector capable of transitioning between flat and line foot configurations while providing grasping capabilities. MAGPIE integrates eight-axis force sensing using proposed mechanisms with Hall effect sensors, enabling both contact and tactile force measurements. We present a computational design framework for our sensing mechanism that accounts for noise and interference, allowing for desired sensitivity and force ranges and generating ideal inverse models. The hardware implementation of MAGPIE is validated through experiments, demonstrating its capability as a foot and verifying the performance of the sensing mechanisms, ideal models, and gated network-based models.
\end{abstract}

\section{Introduction}
The field of limbed robotics, which uses its limbs as both arms and legs, has seen significant growth in recent years thanks to its multi-modal capabilities. %\cite{WAREC_limb}, \cite{shi2021circus}, \cite{hubrobo}, \cite{tanaka2023scaler}, \cite{locomotion_as_manipulation}. %, since they are capable of navigating complex terrains \cite{WAREC_limb}, performing intricate manipulation tasks\cite{shi2021circus}, and climbing while grasping the terrains \cite{hubrobo}.
Limbed robots can interact with objects without a dedicated manipulation arm \cite{shi2021circus}, can traverse various complex terrains \cite{WAREC_limb} and uneven surfaces \cite{locomotion_as_manipulation}, and can climb over obstacles \cite{yusuke_scaler_2022}. 
These abilities expand beyond traditional legged robots' applications, such as search and rescue \cite{WAREC_limb} or planetary exploration \cite{hubrobo}. 

However, the multi-modal nature of limbed robotics presents challenges for end-effector designs since they need to act as both feet for locomotion and grippers for manipulation. 
As a foot, passive and mechanical compliance introduces challenges in legged robot attitude controls since they are not actively controllable \cite{HRP2_ladder}. Contact detection and contact force sensing play a significant role in the legged robot's locomotion \cite{walas2016terrain}. 
Grasping force measurements enable both limb movement and grasping to be force-controlled, allowing for more compliant motions \cite{alex_admittance}, \cite{contact_rich}.
A 6-axis force/torque (F/T) sensor is commonly employed on the ankle, such as in \cite{HRP2_ladder}. However, compact force sensors are essential when the sensor cannot be positioned on the ankle or when tactile information from the foot is required for a specific task \cite{core_shell_ft}. 
The more dynamic capable actuator designs \cite{cheetah}, \cite{zhu2024cycloidal} allow narrow and line feet configurations, which is more suitable for dynamic bipedal robots since the contact model is simplified \cite{hector}.
%Particularly for dynamic bipedal robots, narrow and line feet are more suitable since the contact dynamics and modeling are simplified \cite{hector}. 
However, the line foot design has limited space for contact sensors and is inherently unstable in static and quasi-static cases. 
These challenges motivate a new end-effector design that can realize multiple configurations while achieving high degrees of force sensing. 

% Previous research has explored various gripper and foot mechanisms, including granular jamming feet, monkey feet, and the GOAT gripper. While these offer certain advantages, they often fall short in adaptability, impact resilience, or force sensing capabilities. Specifically, passive mechanisms like magnetic and suction feet cannot dynamically change shape, and active grippers may suffer damage under high-impact conditions.
\begin{figure*}
    \centering
    \includegraphics[width=0.85\textwidth]{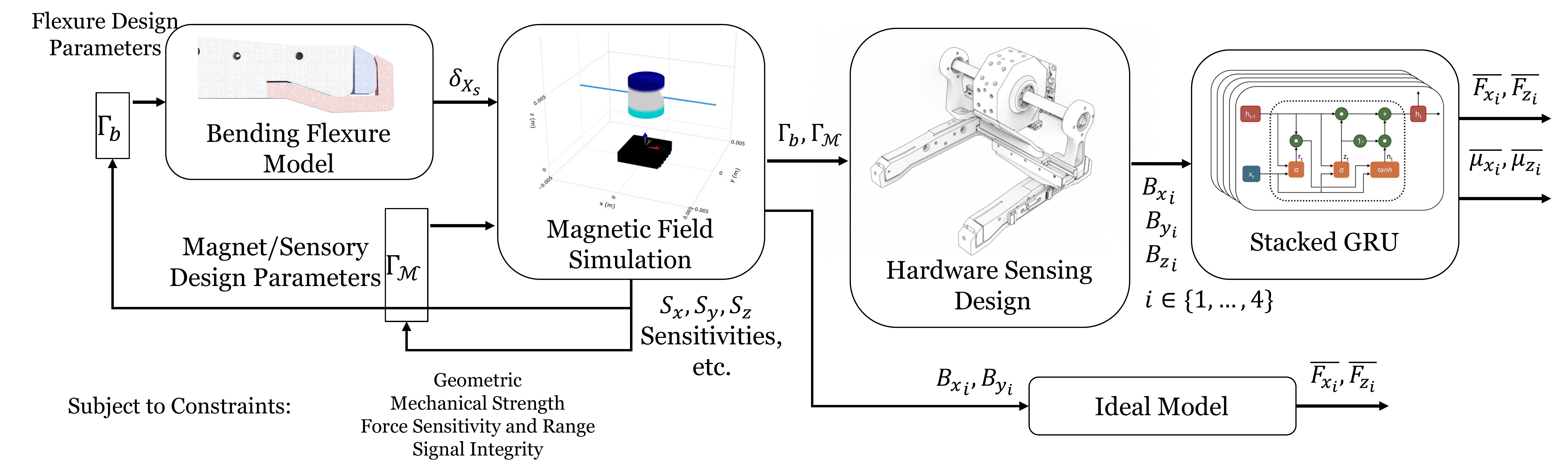}
    \caption{Computational design framework for the Hall effect-based multi-axis force sensing overview. The framework simulated the sensing to design for a desired sensitivity and force range. The framework generates an ideal model using Gaussian radial basis functions. A gated recurrent unit is employed to improve the force measurements and provide uncertainty in the sensing, such as an external significant magnetic field.}
    \label{fig:flow}
\end{figure*}

Hence, we present MAGPIE (Multi-modal Adaptive Gripper for multi-Pedal Impact-resilient End-effector), a two-finger parallel gripper that functions as a multi-modal foot with integrated eight-axis force sensing. MAGPIE can transition between flat and line foot configurations with grasping capabilities, providing adequate modality for limbed robotics that necessitates contacting and grasping end-effectors. 
%The design of sensing mechanisms, accounting for noise, and generating inverse models. 
Our main contributions are:

\begin{itemize}
    \item Developing MAGPIE: a parallel grasping end-effector capable of realizing flat and line foot configurations for limbed robotics.
    \item Computational design framework\footnote{Codes available on https://suke0811.github.io/magpie\_sim/} for 3D Hall effect sensor-based eight-axis force sensing mechanisms. 
    \item verification of hardware MAGPIE design, sensing mechanisms, and the sensing models.
\end{itemize}

\section{Related Works} 
\subsection{Grasping End-Effector as a Foot} 
Gripping and grasping end-effectors have been developed to improve foot traction on challenging terrains, to grasp, adhere, and climb on objects, and to enable locomotion and manipulation with the same end-effector. Adhesion types of grasping mechanisms such as magnetic \cite{magneto}, suction \cite{romein}, and gecko \cite{gecko_grasp_flat} are common mechanisms for legs and grasping. Although these adhesive types are designed for relatively flat surfaces, they could be installed on fingers \cite{stuart2015suction}. 
The granular jamming-based end-effector, conventionally used for grasping objects through its variable compliance \cite{granular_gripper}, has improved the legged robot tractions over various natural terrains \cite{Granular_foot}. This passive adaptation limits control of end-effector geometry, and adding contact sensors is a non-trivial task. 
Multi-fingered grippers with spine \cite{spiny_hand} can grasp objects or terrains, while the spine needles improve microscale contacts with rough surfaces \cite{hubrobo_gripper}, \cite{risk_aware}. However, this type of high degree of freedom (DoF) grippers are less suitable as a foot for dynamic locomotion and are meant for quasi-static locomotion. A multi-modal gripper can grasp a rock and transform it into a wheel \cite{wheel_gripper}. The GOAT gripper \cite{GOAT} has two linear flat fingers and has been effective for climbing and quadruped locomotion, but exhibits structural compliance, and the foot is too narrow to stand stationary. Hence, it was not suitable for bipedal locomotion \cite{scaler-b}.

Therefore, the MAGPIE end-effector aims to achieve grasping capability, while being statically and dynamically suitable as a contacting foot end-effector, such as flat and line foot configurations. The grasping mechanism is designed to be relatively rigid to reduce the control complexity in the limbed robot \cite{HRP2_ladder}.

\subsection{Force, Contact, and Tactile Sensing on end-effectors}
Single and multi-axis F/T sensors are common choices in legged robots \cite{anymal},  \cite{optistate}. These sensors typically consist of strain gauges to measure deflections caused by applied forces \cite{alex_admittance}. However, multi-axis F/T sensors designed for the large force and torque ranges required by adult-sized humanoids are often too large to fit in confined space \cite{core_shell_ft}. Tactile skin sensors, such as arrays of strain gauges or vision cameras \cite{gelslim_iFEM}, are designed for more precise manipulation tasks but tend to be less durable.
Acoustic and viscous flow tactile sensors offer durability and a relatively higher range of force sensing \cite{li2024acoustac}, though the use of audible signals restricts their use case. Multi-modal foot with acoustic, tactile, and capacitive sensors \cite{multi_modal_foot_sense} has been effective in obtaining high dimensional sensing with adaptability for an adult-size bipedal robot, though their focus is on rough terrain classification and locomotion. 
Hall effect sensor-based force sensing mechanisms have been explored for more than 2-axis \cite{hall_effect_tactile}, \cite{hall_effect_3d_force}. However, the multi-axis Hall effect is sensitive and nonlinear \cite{magnetic_joystick_3d}, which requires simulation or empirical tests. 

MAGPIE overcomes these limitations by employing four three-axis Hall effect sensors that detect ground reaction and grasping forces at the edges of each finger, totaling eight-axis force sensing per gripper.
Due to magnetic field sensing's nonlinearity, we introduce a computational design framework to search for various design parameters. This framework also generates an idealized inverse sensing model.

\section{Hardware Design}

\begin{figure}
    \centering
    \includegraphics[width=0.49\textwidth]{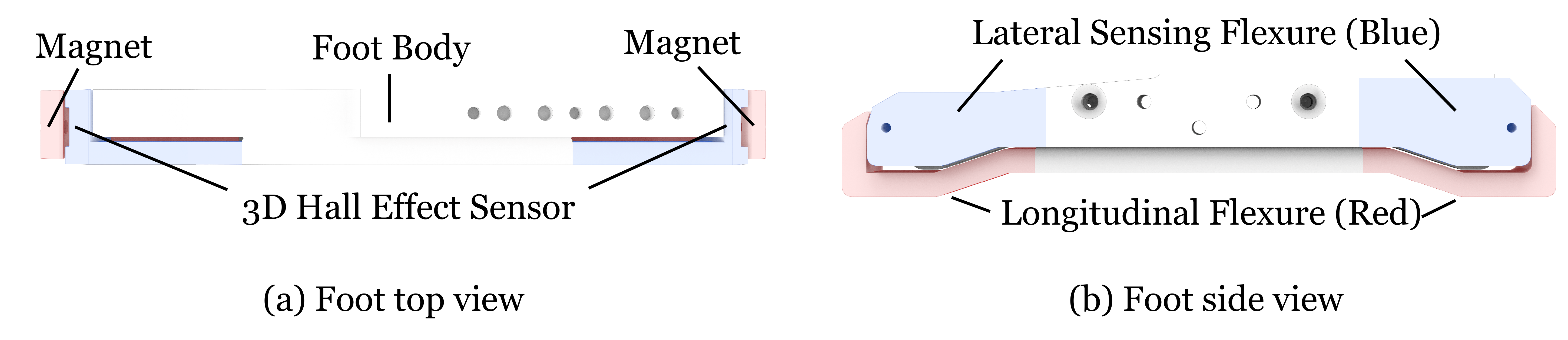}
    \caption{The force sensing mechanism configuration in the hardware. The red and blue flexures are for ground contact and grasping force sensing, respectively. The magnet is attached on the longitudinal, and the sensor is on the lateral side to better isolate the impact forces from the sensor.}
    \label{fig:flexure}
\end{figure}

\begin{figure}
    \centering
    \includegraphics[width=0.49\textwidth]{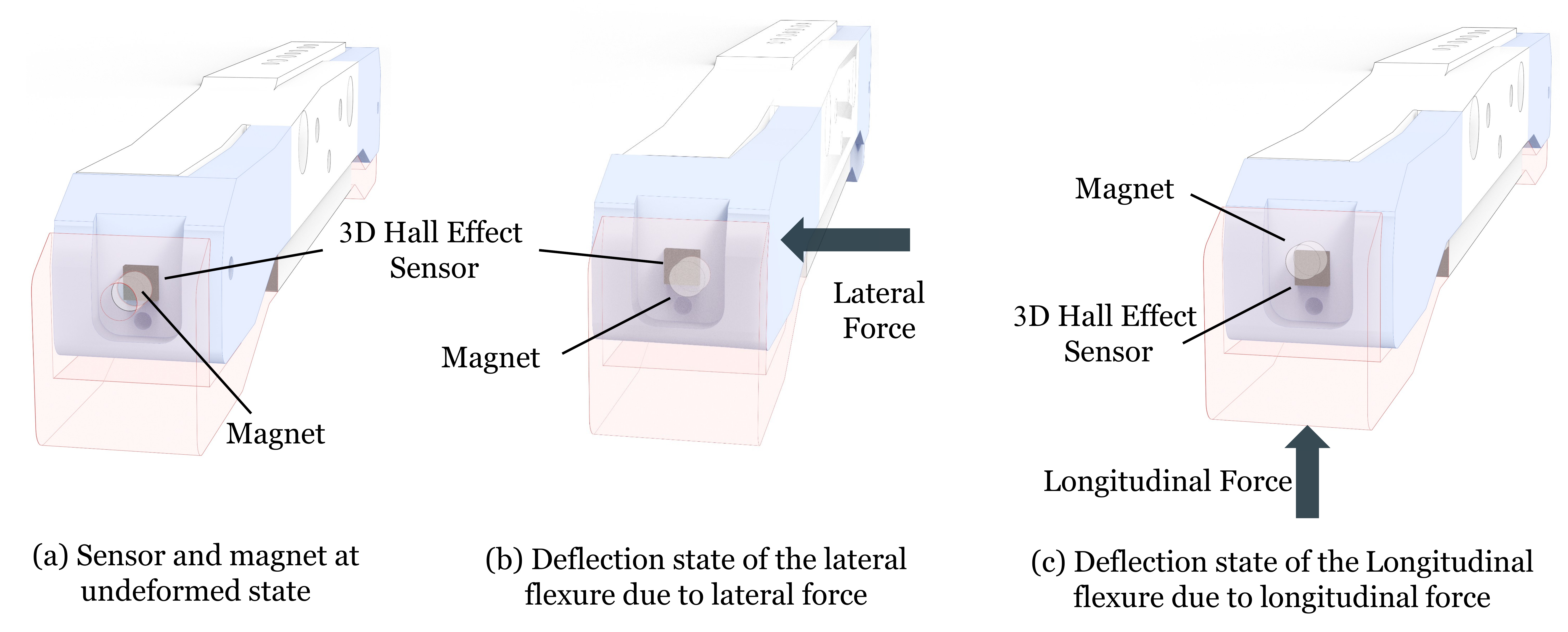}
    \caption{Sensor and magnet placements, and deflection state with forces.}
    \label{fig:deflection}
\end{figure}

\subsection{Foot and Finger Design}

MAGPIE fingers have to be sensitive to both grasping and ground reaction forces. Three-axis Hall effect sensors have been integrated with compliant mechanisms to allow multi-axis force measurements \cite{hall_effect_3d_force}. However, MAGPIE fingers withstand grasping, ground contact, and repeated impact forces. Thus, a 3-axis Hall effect sensor is attached to a one-axial flexure base, and the magnet is installed on the orthogonal axial flexure base as shown in \fig{fig:flexure}. This design avoids having multiple DoF serial motions, increasing the load capacity and allowing independent compliance tuning for each axis. The Hall effect sensor is on the lateral side, i.e., for measuring grasping force, and the magnet is on the longitudinal side, i.e., for measuring ground reaction force. This helps to isolate the electric components and their wires from the landing impact. This design still allows the measurement of two axes simultaneously. \fig{fig:deflection} illustrates the force sensing mechanisms at \fig{fig:flexure}a a nominal state, \fig{fig:flexure}b a deflected state due to lateral force, and \fig{fig:flexure}c a deflected state due to longitudinal force.

% %The lateral and longitudinal compliance are designed to be $xxxx$ mm at $xxxx$ N and $xxxx$ mm at $xxxx$, respectively. 
% The longitudinal flexure may experience unexpectedly greater forces than the sensing range, such as when the robot falls. Hence, flexure includes the hard limit that prevents it from being overloaded. Therefore, the flexure design parameters, $\Gamma_b$, for the flexure include the maximum sensing force, $max(F_{x})$, $max(F_{y})$, bending deflection range, $\delta w$, the maximum survival force, $max(F_{x})$, $max(F_{y})$. 

\subsection{The Gripper Actuation Mechanisms \label{sec:actuation_mech}}
The MAGPIE's actuation mechanism has to withstand unconventional longitudinal stress due to the ground contact. Hence, we opt for a miniature crossed-roller linear rail system actuated with a BLDC motor. 
The lateral force sensing flexure is designed to be more sensitive, but once the gripper is closed, the lateral sensing mechanism is not exposed and hence protected. The right and left-handed lead screws are connected at the BLDC inner rotor using MechaLock to distribute torque transfer stress evenly.

\subsection{Circuitry and Sensors}
\begin{figure}
    \centering
    \includegraphics[width=0.32\textwidth]{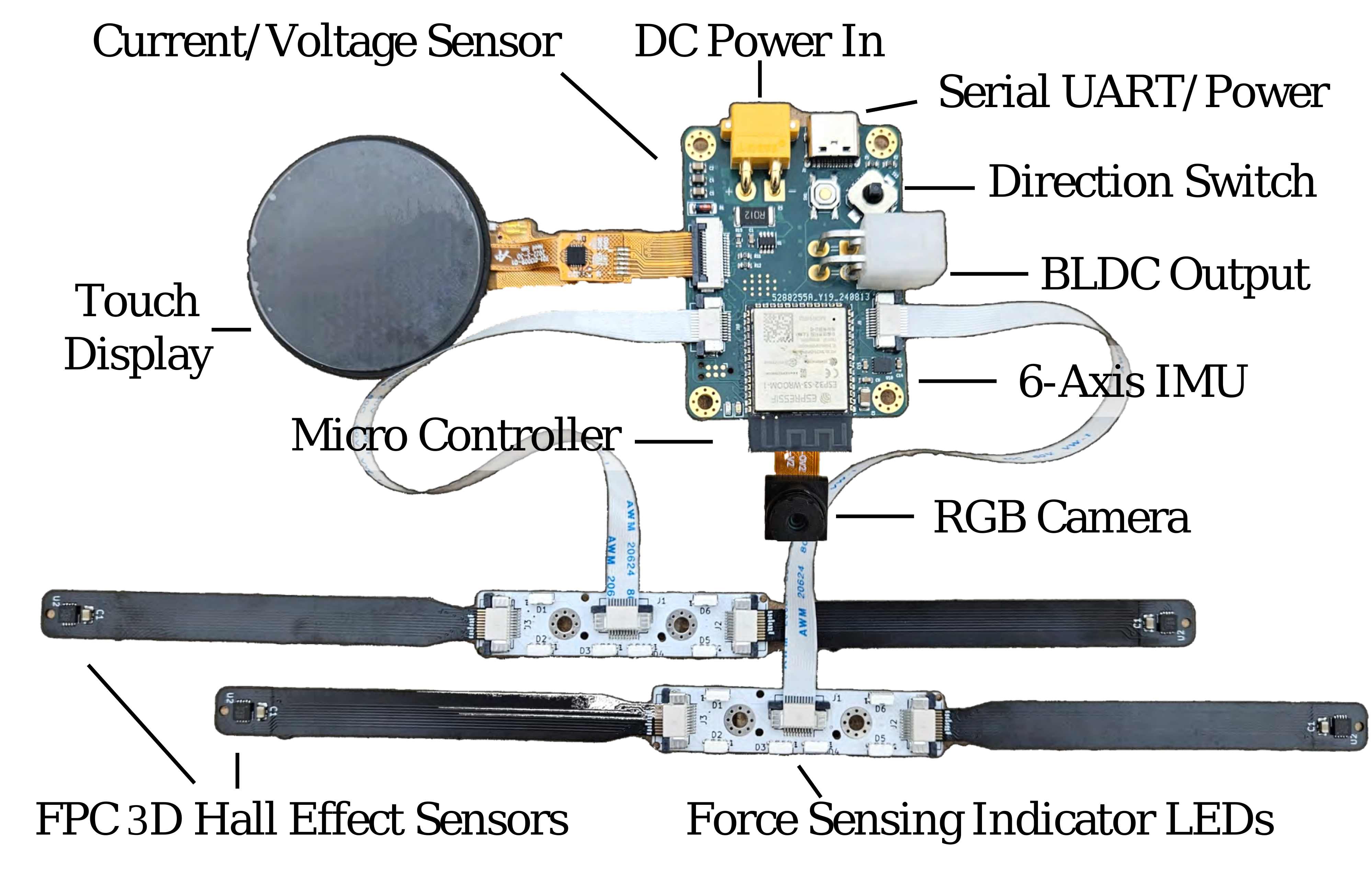}
    \caption{Circuitry and sensors of the MAGPIE control unit.}
    \label{fig:pcb}
\end{figure}

The electronics of the MAGPIE gripper, shown in \fig{fig:pcb}, are designed to be modular and self-contained, allowing for straightforward integration with the rest of the robotic system. The electrical circuit boards are developed using Polymorphic Blocks \cite{PolymorphicBlocks} and its human-computer interface \cite{PolymorphicBlocksEdge}, which enables schematic, component-level design, electrical modeling, and design verification.
The MAGPIE PCB board integrates a BLDC motor, a 6-axis inertial measurement unit (IMU), an RGB camera, a current/voltage sensor, a color touch display, DC-DC converters, and interfaces via UART and CANBus. The IMU and the cameras will benefit the future end-effector state estimation and visual servo.  
The 3-axis Hall effect sensor is soldered on a custom flexible PCB to minimize wiring and simplify sensor placement, reducing potential points of failure.
% and reduce hardware complexity. This flexible PCB design minimizes wiring and , 

\section{Sensor Model and Design Framework}
The flexure design, magnet characteristics, 3D Hall effect sensors, and magnet placements are vital in MAGPIE contact and tactile force sensing. Hence, these designs require a design framework to determine the appropriate design parameters to achieve proper force sensing range, sensitivity, repeatability, and insensitivity to disturbance. 

\subsection{Flexure Analytical and Simulation Model}

Here, we employ Von Kármán beam theory to estimate the flexure responses to external forces analytically. While the Euler-Bernoulli beam theory is valid for small deformations, capturing both horizontal and vertical deformations in this design is essential, as the sensor is sufficiently sensitive to detect these effects. The Von Kármán beam theory is suited for describing the large deflections of slender beams by accounting for both bending and in-plane stretching.
For the vertical force balance (bending equation), we have:

\begin{equation}\label{eq:vertical}
    EI \frac{d^4 w(x)}{dx^4} = q(x) - \frac{d}{dx} \left( N(x) \frac{dw(x)}{dx} \right)
\end{equation}

Where $E$ is Young’s modulus, and $I$ is the second moment of the area of the beam. $w(x)$ denotes the vertical beam deflection at a position $x$ along its length. $N(x)$ is the beam axial force, and $q(x)$ represents the distributed transverse load per unit length.

For the horizontal force balance (stretching equation):

\begin{equation} \label{eq:horizontal}
\frac{d}{dx} \left( EA \frac{du(x)}{dx} \right) = \frac{1}{2} EI \left( \frac{dw(x)}{dx} \right)^2
\end{equation}

Here, $u(x)$ is the horizontal displacement (in-plane stretching) at position $x$, and $A$ is the beam's cross-sectional area.

% When the axial force $N(x)$ is negligible, \eq{eq:vertical} simplifies bending equation as:

% \begin{equation} \label{eq:vertical_simple}
% EI \frac{d^4 w(x)}{dx^4} = P \cdot \frac{L - x}{L}
% \end{equation}

% The beam edge's angle is derived from the slope of the deflection curve, $w(x)$. 
The derivative, $\frac{dw(x)}{dx}$, gives the slope at any point along the beam.
Using the expression for the second derivative of $w(x)$, the slope at the edge can be obtained as:
\begin{equation} 
\theta_{\text{edge}} = \frac{dw(L)}{dx} = \int_0^L \frac{P \cdot (L - x)}{EI} \, dx
\end{equation}

\subsection{Magnetic Sensor Signal Model}
A computational model and estimate of the sensor signal are vital steps in designing a desired Hall effect sensor-based force sensing system. 
Sensor signal patterns differ based on the magnet's magnetization strength, polarity, sizes, and relative placement with respect to the sensor. 
Here, we consider cylindrical magnets due to their wider size availability, and the cube or spherical magnets do not affect the signal patterns at micron-scale motion sensing as shown in \fig{fig:magnet_two}a. 
The magnetic field change sensitivity, $\mathcal{S}$, is a rate of the magnetic field change at the sensor frame due to the magnet motion, denoted as $\mathcal{S} = \frac{\delta \mathcal{B}}{\delta X}$. The higher sensitivity can translate to the relatively minor deflection of the beam to generate a larger change in the magnetic field at the sensor. This allows us to design the sensing mechanisms with appropriate force resolution, although the Hall effect sensors' sensitivity is set. The magnetic field numerical simulation is implemented using Magpylib \cite{magpylib}.

\subsection{Magnetic Interference and Uncertainty\label{sec:interference_model}}
The Hall effect sensor-based force sensing is sensitive to magnetic interference.
%, such as 1) Earth's magnetic field, 2) sensor and magnet misalignment, 3) nearby permanent magnet, i.g., another sensor magnet when the gripper is closed, and 4) permanent magnet inhomogeneity. 
Conventionally, a strong permanent magnet is simply employed to convey such interference \cite{hall_effect_tactile}.
%the external magnetic interference can be conveyed by having a significantly stronger permanent magnet for the sensor \cite{hall_effect_tactile}. 
Here, we incorporate interference models to determine the design parameters, changing the mechanism design parameters to mitigate the expected disturbances. 

\subsubsection{Earth's magnetic field}
The Earth's magnetic field causes background noise, which depends on the orientation of the sensors. However, this field strength is approximately $0.5$ $\mu$T, whereas N32 grade NdFeB magnets are $1.2$ T. Hence, the effect is negligible where the sensor is placed in proximity to the magnet. 

\subsubsection{Sensor and magnet misalignment}
The sensor and magnet placements have manufacturing uncertainty, which causes biases and asymmetry in the sensor measurements. Although the biases are relatively more straightforward to calibrate, the asymmetry introduces skew in sensor measurements, albeit the measurement-to-force model becomes distinct for each axis.

% \subsubsection{Electromagnetic Interference}
% The BLDC leakage is dynamically variable depending on the BLDC dynamics, requiring multi-physics FEA simulation \cite{zhang2017leakage}. Our framework simplifies BLDC and radial coil with changing phase current.  
% The external leakage field generated by each stator coil can be estimated using the Biot-Savart law, which calculates the magnetic field at a point due to a current-carrying conductor as: 

% \begin{equation}
%     B_\kappa = \frac{\mu_0}{4\pi} \cdot \frac{n I_p \, l \times \hat{r}}{r^2}
% \end{equation}

% Where \( \mu_0 \) is the permeability of free space, \( n \) is the number of turns in the stator coil, \( I_p \) is the phase current flowing through the windings, \( l \) is the coil’s length, \( r \) is the distance between the point of interest (e.g., where the sensor is located) and the current element, and \( \hat{r} \) is the unit vector pointing from the current element to the point of interest. The coil is radially allocated, but due to the multi-phase nature of BLDC, they do not cancel out the leakage field. 
% The rotor permanent magnets can be considered as well in the same way as the sensor magnet. However, since MAGPIE uses an inner rotor BLDC, the leakage is dominant due to the coil. 

\subsubsection{Permanent magnet nearby}
In MAGPIE, another permanent magnet can be nearby when the gripper is closed. Instead of shielding or compensating for this interference,
%Although this could be tried to shield or compensate by knowing the gripper finger current distance, 
our computational framework allows us to design sensing mechanisms that do not affect the measurements more than the desired level at most. 

\subsubsection{Permanent magnet impurity}
Demagnetization effects in permanent magnets arise from the inhomogeneous response of the magnetic material, leading to a slight weakening of the magnetic field. For NdFeB (neodymium) magnets, the demagnetization effect is typically less than $1\%$ \cite{magnetic_joystick_3d}. 

\begin{figure*}
    \centering
    \includegraphics[width=0.99\textwidth, trim={0.24cm 0.75cm 2cm 0.75cm}, clip]{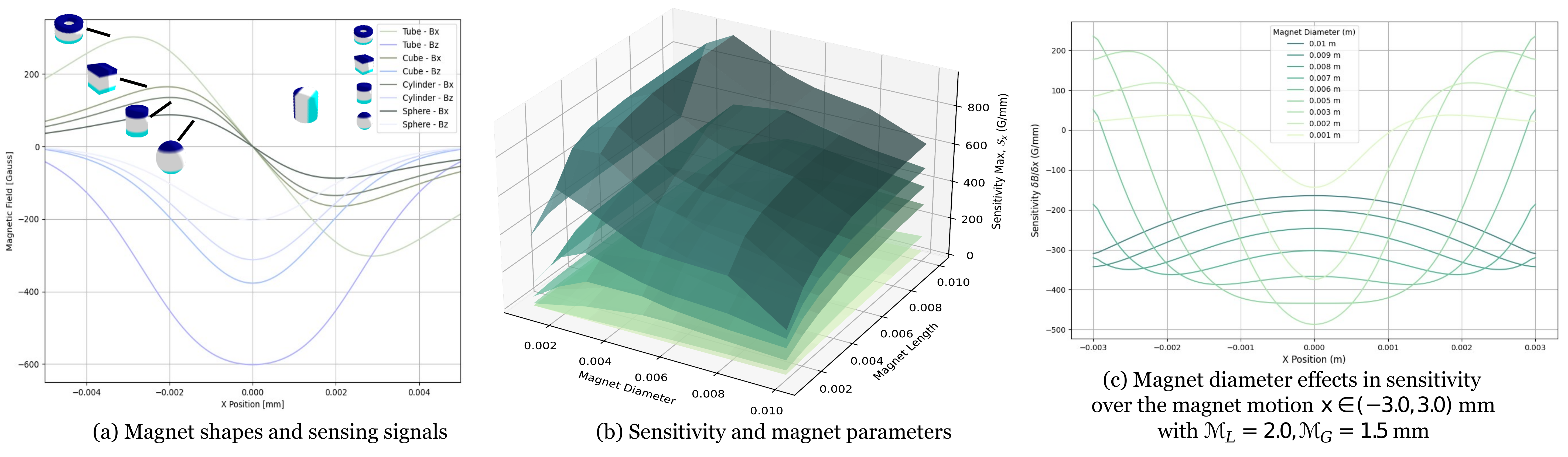}
    \caption{Simulated magnetic field changes at the sensor frame. The magnets purely move in the $x$ direction. For axial metric magnets, the motion in the $x$ and $y$-axis generates the identical signal.}
    \label{fig:magnet_two}
\end{figure*}

\subsection{Constraints\label{sec:constraints}}
The sensing mechanism should prevent bending yield stress and fatigue failure. The maximum bending stress is:

\begin{equation} \sigma_{b}(x) = \frac{M(x) \cdot c}{I} \end{equation}

Where $M(x)$ is the bending moment at position $x$, and $c$ is the distance from the neutral axis to the outermost fiber.
%, and $I$ is the second moment of area. 
To avoid fatigue failure, the stress amplitude $\sigma_a = \frac{\sigma_{\text{max}} - \sigma_{\text{min}}}{2}$ should correspond to an indefinite fatigue life $N_f$ according to the material's S-N curve.

In addition to geometric constraints due to the sensor and magnet sizes, the force-sensing range can be limited by having a collision under excessive forces. The sensor survival force range can be increased mechanically by limiting the maximum bending deflection.
The estimated magnetic field strength at the sensor must be within the 3D Hall effect sensor range. The possible magnetic disturbances discussed in \ref{sec:interference_model} should be considered as well.

\subsection{Force Sensing Model\label{sec:model}}
Estimating contact and grasping forces from 3D Hall effect sensor measurements requires inverting the analytical model. Since deriving a closed-form inverse is challenging, numerical and data-driven methods are used in multi-axis force estimation, including inverse FEA \cite{gelslim_iFEM} and neural networks \cite{hall_effect_tactile}. 
We generate an ideal model based on simulated sensor measurements for given applied forces, as illustrated in \fig{fig:flow}, and employ Gaussian radial basis functions (GRBF) \cite{grbf} to model the inverse relationship.

We incorporate real-world data from the MAGPIE hardware to account for uncertainties, such as sensor nonlinearity and hysteresis, not considered in our analytical models. Given the time-dependent nature of hysteresis, we select a stacked Gated Recurrent Unit (GRU) \cite{chung2014empiricalevaluationgatedrecurrent}, which takes sensor measurements as inputs and outputs estimated mean forces and their uncertainties. Estimating uncertainty is vital to detect unmodeled magnetic field disturbances, such as external magnets, since significant uncertainty can indicate abnormal operating conditions.

\section{Results and Hardware Experiments}
In this section, we conduct 1) the 3D Hall effect sensor-based force sensing mechanism design parameter search through our computational framework. 2) the sensing mechanisms, ideal, and GRU-based sensing model verification and analysis. 3) Testing of the MAGPIE as a gripper and foot.
\subsection{Magnet Shapes and Sizes}
\fig{fig:magnet_two}a shows the magnetic field changes at the sensor for four different magnet shapes. All magnet sizes have a constant minimum bounding box of $2.5$ mm, and the tube has a hole diameter of $1$ mm. The magnetic field strength varies based on the shape due to differences in the volume.

\subsection{Sensitivity, Magnet Parameters with Self-Interference\label{sec:sesor_sensitivity}}
Here, using the computational design framework, we simulate the relationship among the maximum sensitivity, $\max(\mathcal{S})$ magnet sizes, and the relative distance to the Hall effect sensor. The magnet is moved along with the $x$ axis of the sensor in the range of $(-3.0, 3.0)$ mm, and the simulation results in \fig{fig:magnet_two}b. The identical size of the magnet is placed at $1.5$ mm to represent the case where the gripper is closed, i.e., the closest distance to the other foot magnet. 

The maximum sensitivity is proportional to the distance. The longer magnet increases the sensitivity in a log-function manner. The larger diameter magnets do not necessarily increase the sensitivity. \fig{fig:magnet_two}c indicates that the sensitivity changes significantly over this magnet $x$ motion. $\mathcal{M}_D=5.0$ mm exhibits nearly constant sensitivity for $x \in (-0.6, 0.6)$ mm. The larger diameter magnets add nonlinearity in $\mathcal{S}_x$ around $x=0$ mm. 
The disturbance due to the neighboring magnets leaked flux in \fig{fig:magnet_two}b increases corresponding to $\mathcal{M}_d$ and $\mathcal{M}_L$. The $\delta B_{x_e}$ will constantly offset the Hall effect sensor measurement given the gripper open or close states.

% \begin{figure}
%     \centering
%     \includegraphics[width=0.4\textwidth, trim={1cm 0cm 1cm 1cm}, clip]{fig_magpie/sensitivity.jpeg}
%     \caption{The sensitivity, $\mathcal{S}_x$ changes over the magnet motion $x\in (-3.0, 3.0)$ mm with $\mathcal{M}_L = 2.0$ mm and $\mathcal{M}_G = 1.5$ mm.}.
%     \label{fig:magnet_sense}
% \end{figure}

\begin{figure}
    \centering
    \includegraphics[width=0.48\textwidth, trim={0.2cm 1.cm 0cm 0.8cm}, clip]{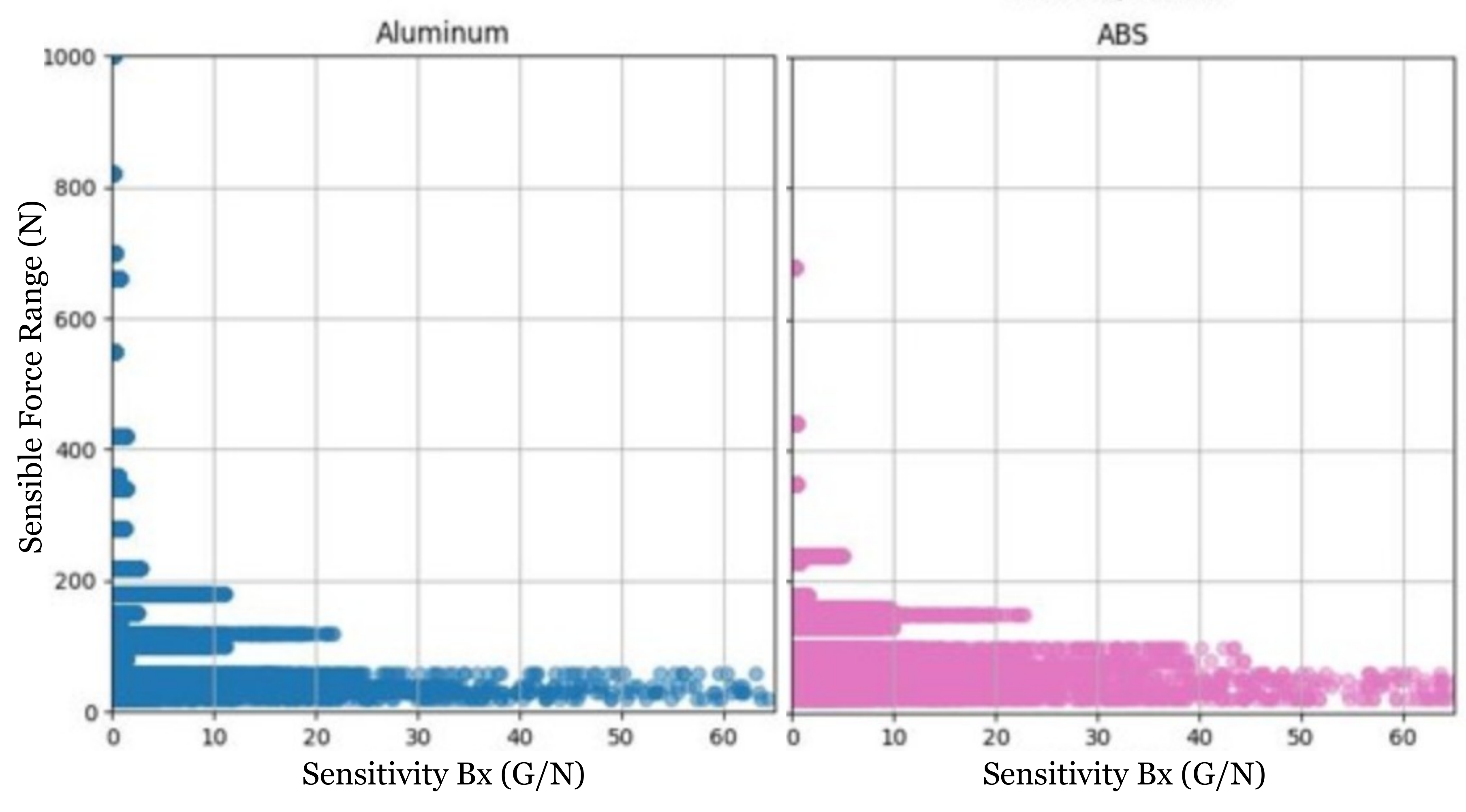}
    \caption{Parameter sweep results of $10000$ design parameters for two different materials, showing one metal and plastic material result. }
    \label{fig:material}
\end{figure}

\subsubsection{Sensing Mechanisms Design Search}
Using the entire computational design pipeline, here we analyze the input-output relationships of the applied force and the 3D Hall effect sensing with various $\Gamma_b$, $\Gamma_{\mathcal{M}}$ parameters, such as materials and magnet sizing. \fig{fig:material} plots the possible force sensitivity and the force ranges for different materials. The force sensitivity is affected by both $\mathcal{S}$ in Section. \ref{sec:sesor_sensitivity} and the stiffness of the beam. The force range achievable is constrained by the fatigue and bending failure as described in Section. \ref{sec:constraints}. \fig{fig:material} indicates the suitable materials for users' desired force ranges and sensitivity. Both metal and plastic can achieve similar sensing ranges and sensitivity at low ranges $(0, 10)$ Gauss/N, $(0, 100)$ N. In Plastic materials such as ABS, the maximum force range is $(0, 200)$ N. On the other hand, the metal can achieve the force range of $800$ N, but the sensitivity is less than $\mathcal{S}_x=10$ Gauss/N. Parameter sweeping range and resolutions are limited, resulting in sparse distribution for the higher force range in \fig{fig:material}

\begin{figure*}
    \centering
    \includegraphics[width=0.9\textwidth,trim={0.5cm 0cm 0cm 0cm},clip]{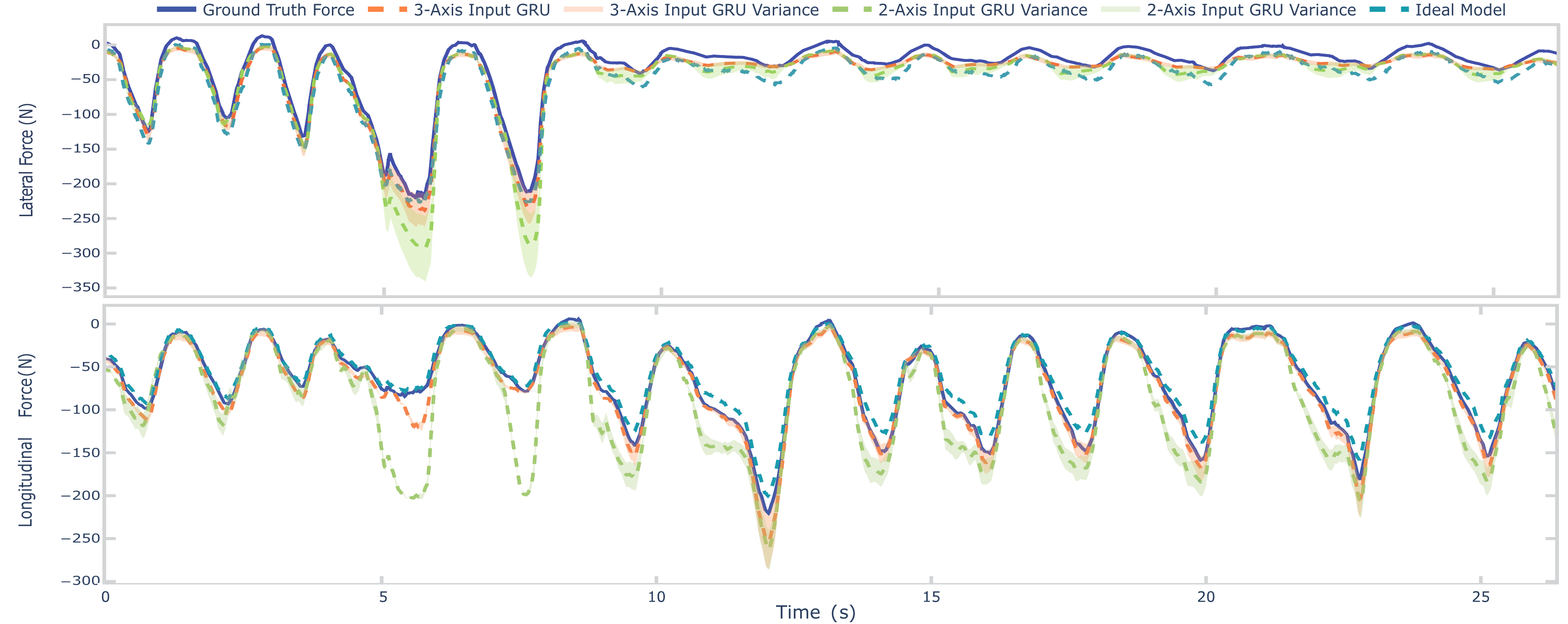}
    \caption{The ground truth, the ideal model, and the stacked GRU with 2-axis and 3-axis inputs results from hardware.}
    \label{fig:gru_compare}
\end{figure*}

\begin{figure}
    \centering
    \includegraphics[width=0.49\textwidth,trim={0cm 0cm 0cm 0cm},clip]{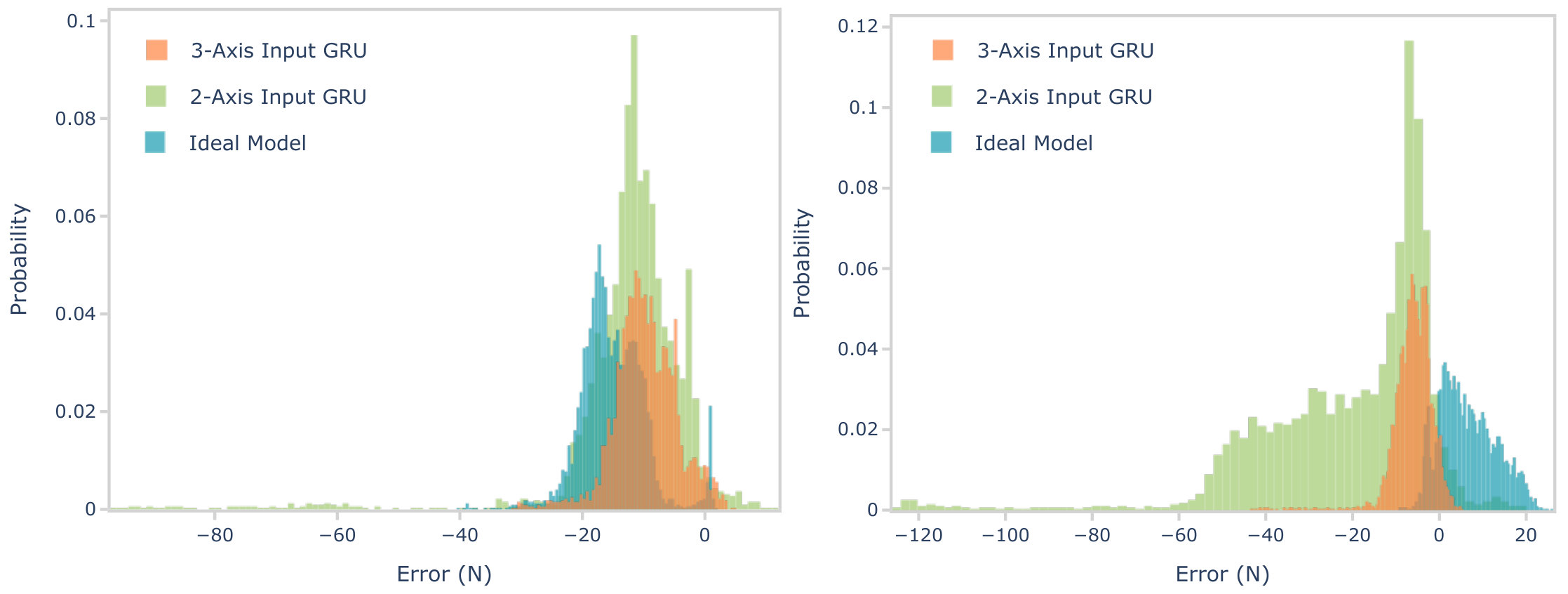}
    \caption{Model error distribution for the data in \fig{fig:gru_compare}.}
    \label{fig:gru_compare_dist}
\end{figure}

\subsubsection{Design Selection}
Hardware MAGPIE employs A31301 3D linear hall-effect sensors by ALLEGRO Microsystems with a $\pm 2000$ Gauss range and $0.1$ Gauss of resolution, which comes with internal compensation such as temperature effects. The design search set the beam deflection range to $0.5$ mm, and the flexure beam was manufactured using wire EDM. The effective beam length, thickness, width were $30$, $5$ and $15$ mm, respectively
Due to the sensor packages and structure limitation, the sensor and the magnet distance $\mathcal{M}_G$ had to be at least $1.5$ mm. From \fig{fig:magnet_two}c, The magnet $\mathcal{M}_D=0.003$ m was selected for the highest sensitivity. The sensitivity is nonlinear for the sensing range, which is complex in the coaxial force cases and is handled through a stacked GRU framework. 
When the gripper is closed, this mechanism's configuration has less than $\pm 3$ resolutions of A31301 from the simulation.

\subsection{GRBF and GRU Model on Hardware}
\subsubsection{Data Collection}
The ground truth force was recorded with two Bota Rokubi F/T sensors attached to each flexure axis at $100$ Hz to determine the relationship between the force applied and the magnetic sensor measurements. Each F/T and Hall effect sensor was bias-calibrated to account for the misalignment and variances of the magnet. All four Hall effect sensors run at $1000$ Hz on board.
The GRBF ideal model force estimate is from the simulation in Section. \ref{sec:model}. \fig{fig:gru_compare} compares the ground truth applied forces, idea model estimated forces, and the stacked GRU estimated forces with $2$-axis and 3-axis sensor inputs. Each model estimation root square mean error (RSME) and statistical results are listed in Table. \ref{tb:compare}. 
A sinusoidal lateral $F_x$ and longitudinal $F_z$ force is applied with varying amplitude with an average frequency of $0.5$ Hz as shown in \fig{fig:gru_compare}.  The error normalized histogram in \fig{fig:gru_compare_dist} was obtained over the $100$ s test. 

\subsubsection{Statistical Analysis and discussion}
The GRBF ideal model has shown statistical relationships to the hardware sensor-force correlation but underestimates the force at high amplitude. 
All models showed no statistically meaningful differences in the phase delay in estimation except for 2-Axis GRU $F_z$. The 2-axis GRU model struggled particularly when both axes experienced significant deviation, such as at $5.7$ and $7.7$ s. The 3-axis GRU has handled the coupled cases better than the 2-axis GRU with the same amount of data due to additional redundant information indicating the coupling of both axes. The uncertainty estimation increases as the 3-axis GRU deviates from the ground truth. 

The 3-axis GRU model outperforms the ideal model, the 2-axis GRU model. However, the ideal model estimation was consistently closer to the ground truth around $<25$ N or $<10$ \% of the force range. This is potential because the lower force range dataset is more sensitive to the F/T and Hall effect sensors' initial calibration. In such cases, GRU requires more data to learn biases. In the future, adding the ideal model estimation as GRU input can enhance the training data efficiency and improve the estimation quality since the ideal model works as a warm start initial guess. The uncertainty estimation around zero can be potentially improved with non-uniform normalization to force learning in these regions \cite{dnn_normalize}.

\subsubsection{Computation Cost and External Interference}
There were no statistically meaningful deviations in the GRU force estimation due to known magnetic interference, such as from BLDC and when the gripper is closed. 
The ideal model runs at $11$ us, and the GRU takes $22$ us on CPU and $15$ us on GPU, including data transfer time. The ideal model can be faster if a linear model is used instead. Where an external magnetic field is applied, the GRU uncertainty output consistently increases up to $\pm 50$ N, which can be used to identify the abnormality. This is another essential motivation for adopting GRU instead of the GRBF ideal model, which only provides the mean estimation.

\begin{table}[t!]
 \centering
 \caption{Force sensing errors compared to the ground truth.}
\label{tb:compare}
\begin{tabular}{c|ccc}
 & Ideal Model & GRU 2Axis & GRU 3Axis \\ \hline
 $F_z$ RMSE [N] & 9.2 & 28.9 & 7.6 \\
\rowcolor[HTML]{EFEFEF}
$F_z$ Mean Error [N] & 6.7 & -20.4 & -5.9 \\
$F_z$ Variance [$N^2$] & 6.4 & 20.5 & 4.7\\
$F_x$ RMSE [N] & 16.0 & 15.8 & 11.0 \\
\rowcolor[HTML]{EFEFEF}
$F_x$ Mean Error [N] & -15.1 & -11.9 & -9.8 \\
$F_x$ Variance [$N^2$] & 5.4 & 10.4 & 5.1 
\end{tabular}
\end{table}

\subsection{Grasping and Foot Capability}
Here, we verify MAGPIE's grasping and capability as a foot. MAGPIE employs a custom direct drive BLDC with rated speed and torque of $114$ rad/sec and $0.58$ N/m and peak torque of $0.96$ N/m. With the lead screw mechanisms, the nominal gripper opening and closing speed was $58.3$ mm/sec, and the nominal grasping force was $350$ N, which was measured using our Hall effect sensors. 
The rigidity of the entire mechanism was evaluated by applying a static load of $200$ N at the top of the gripper and measuring the deflection, which was $0.3$ mm.

\section{Conclusion}
This paper presented MAGPIE, an end-effector capable of grasping an object while functioning as both line and flat feet with contact/tactile force sensing through 3D Hall effect sensors. Our computational framework enables the design of the sensing mechanisms, including flexure beams, magnet, and sensor parameters, while considering nonlinear effects and magnetic disturbances.
%, such as interference from neighboring sensor magnets when the gripper is closed.
MAGPIE mechanisms and force-sensing capabilities have been verified on the hardware using an ideal GRBF model generated from the framework and stacked GRU. MAGPIE enables the development and deployment of limbed robotic systems that can leverage MAGPIE's multi-modality and total eight-axis contact and tactile force sensing for conducting, such as simultaneous locomotion, manipulation, and grasping tasks.
%Our computational design framework provides a way to further develop Hall effect-based force sensing mechanisms that are more resilient and less prone to self-interference. 

Our multi-axis sensing mechanism can be extended and applied to smaller or larger end effector force sensing scenarios thanks to the simplicity of the mechanism and scalability provided by the computational framework. MAGPIE computational framework provides users "off-the-shelf" ideal models for customized sensing mechanisms, while allowing users to tailor the sensor model with our GRU framework once data is collected. Hence, our work will aid the broader adoption of multi-axis Hall effect sensor-based force sensing. 
MAGPIE showcased multi-modal robotic end-effectors, promising enhanced versatility in multi-modal robotics.

\newpage

\bibliographystyle{IEEEtran}
\bibliography{main}

\begin{thebibliography}{10}
\providecommand{\url}[1]{#1}
\csname url@samestyle\endcsname
\providecommand{\newblock}{\relax}
\providecommand{\bibinfo}[2]{#2}
\providecommand{\BIBentrySTDinterwordspacing}{\spaceskip=0pt\relax}
\providecommand{\BIBentryALTinterwordstretchfactor}{4}
\providecommand{\BIBentryALTinterwordspacing}{\spaceskip=\fontdimen2\font plus
\BIBentryALTinterwordstretchfactor\fontdimen3\font minus \fontdimen4\font\relax}
\providecommand{\BIBforeignlanguage}[2]{{%
\expandafter\ifx\csname l@#1\endcsname\relax
\typeout{** WARNING: IEEEtran.bst: No hyphenation pattern has been}%
\typeout{** loaded for the language `#1'. Using the pattern for}%
\typeout{** the default language instead.}%
\else
\language=\csname l@#1\endcsname
\fi
#2}}
\providecommand{\BIBdecl}{\relax}
\BIBdecl

\bibitem{shi2021circus}
F.~Shi, T.~Homberger, J.~Lee, T.~Miki, M.~Zhao, F.~Farshidian, K.~Okada, M.~Inaba, and M.~Hutter, ``Circus anymal: A quadruped learning dexterous manipulation with its limbs,'' in \emph{2021 IEEE International Conference on Robotics and Automation}.\hskip 1em plus 0.5em minus 0.4em\relax IEEE, 2021, pp. 2316--2323.

\bibitem{WAREC_limb}
K.~Hashimoto, S.~Kimura, N.~Sakai, S.~Hamamoto, A.~Koizumi, X.~Sun, T.~Matsuzawa, T.~Teramachi, Y.~Yoshida, A.~Imai \emph{et~al.}, ``Warec-1—a four-limbed robot having high locomotion ability with versatility in locomotion styles,'' in \emph{2017 IEEE International Symposium on Safety, Security and Rescue Robotics (SSRR)}.\hskip 1em plus 0.5em minus 0.4em\relax IEEE, 2017, pp. 172--178.

\bibitem{locomotion_as_manipulation}
T.~G. Chen, S.~Newdick, J.~Di, C.~Bosio, N.~Ongole, M.~Lapôtre, M.~Pavone, and M.~R. Cutkosky, ``Locomotion as manipulation with reachbot,'' \emph{Science Robotics}, vol.~9, no.~89, p. eadi9762, 2024.

\bibitem{yusuke_scaler_2022}
Y.~Tanaka, Y.~Shirai, X.~Lin, A.~Schperberg, H.~Kato, A.~Swerdlow, N.~Kumagai, and D.~Hong, ``Scaler: A tough versatile quadruped free-climber robot,'' in \emph{2022 IEEE/RSJ International Conference on Intelligent Robots and Systems}, 2022, pp. 5632--5639.

\bibitem{hubrobo}
K.~Uno, N.~Takada, T.~Okawara, K.~Haji, A.~Candalot, W.~F.~R. Ribeiro, K.~Nagaoka, and K.~Yoshida, ``Hubrobo: A lightweight multi-limbed climbing robot for exploration in challenging terrain,'' in \emph{2020 IEEE-RAS 20th International Conference on Humanoid Robots}, 2021, pp. 209--215.

\bibitem{HRP2_ladder}
J.~Vaillant, A.~Kheddar, H.~Audren, F.~Keith, S.~Brossette, K.~Kaneko, M.~Morisawa, E.~Yoshida, and F.~Kanehiro, ``Vertical ladder climbing by the hrp-2 humanoid robot,'' in \emph{2014 IEEE-RAS International Conference on Humanoid Robots}, 2014, pp. 671--676.

\bibitem{walas2016terrain}
K.~Walas, D.~Kanoulas, and P.~Kryczka, ``Terrain classification and locomotion parameters adaptation for humanoid robots using force/torque sensing,'' in \emph{2016 IEEE-RAS 16th International Conference on Humanoid Robots (Humanoids)}.\hskip 1em plus 0.5em minus 0.4em\relax IEEE, 2016, pp. 133--140.

\bibitem{alex_admittance}
A.~Schperberg, Y.~Shirai, X.~Lin, Y.~Tanaka, and D.~Hong, ``Adaptive force controller for contact-rich robotic systems using an unscented kalman filter,'' in \emph{2023 IEEE-RAS 23rd International Conference on Humanoid Robots}, 2023.

\bibitem{contact_rich}
Y.~Shirai, X.~Lin, A.~Schperberg, Y.~Tanaka, H.~Kato, V.~Vichathorn, and D.~Hong, ``Simultaneous contact-rich grasping and locomotion via distributed optimization enabling free-climbing for multi-limbed robots,'' in \emph{2022 IEEE/RSJ International Conference on Intelligent Robots and Systems}.\hskip 1em plus 0.5em minus 0.4em\relax IEEE, 2022, pp. 13\,563--13\,570.

\bibitem{core_shell_ft}
K.~Shinjo, K.~Kawaharaduka, Y.~Asano, S.~Nakashima, S.~Makino, M.~Onitsuka, K.~Tsuzuki, K.~Okada, K.~Kawasaki, and M.~Inaba, ``Foot with a core-shell structural six-axis force sensor for pedal depressing and recovering from foot slipping during pedal pushing toward autonomous driving by humanoids,'' in \emph{2019 IEEE/RSJ International Conference on Intelligent Robots and Systems (IROS)}, 2019, pp. 3049--3054.

\bibitem{cheetah}
S.~Seok, A.~Wang, M.~Y. Chuah, D.~Otten, J.~Lang, and S.~Kim, ``Design principles for highly efficient quadrupeds and implementation on the mit cheetah robot,'' in \emph{2013 IEEE International Conference on Robotics and Automation}.\hskip 1em plus 0.5em minus 0.4em\relax IEEE, 2013, pp. 3307--3312.

\bibitem{zhu2024cycloidal}
A.~Zhu, Y.~Tanaka, F.~Rafeedi, and D.~Hong, ``Cycloidal quasi-direct drive actuator designs with learning-based torque estimation for legged robotics,'' in \emph{2025 IEEE International Conference on Robotics and Automation (ICRA)}, 2025.

\bibitem{hector}
J.~Li, J.~Ma, O.~Kolt, M.~Shah, and Q.~Nguyen, ``Dynamic loco-manipulation on hector: Humanoid for enhanced control and open-source research,'' \emph{arXiv preprint arXiv:2312.11868}, 2023.

\bibitem{magneto}
T.~Bandyopadhyay, R.~Steindl, F.~Talbot, N.~Kottege, R.~Dungavell, B.~Wood, J.~Barker, K.~Hoehn, and A.~Elfes, ``Magneto: A versatile multi-limbed inspection robot,'' in \emph{2018 IEEE/RSJ International Conference on Intelligent Robots and Systems (IROS)}, 2018, pp. 2253--2260.

\bibitem{romein}
C.~Prados, M.~Hernando, E.~Gambao, and A.~Brunete, ``Moclora \&mdash;an architecture for legged-and-climbing modular bio-inspired robotic organism,'' \emph{Biomimetics}, vol.~8, no.~1, 2023.

\bibitem{gecko_grasp_flat}
D.~Hirano, N.~Tanishima, A.~Bylard, and T.~G. Chen, ``Underactuated gecko adhesive gripper for simple and versatile grasp,'' in \emph{2020 IEEE International Conference on Robotics and Automation (ICRA)}, 2020, pp. 8964--8969.

\bibitem{stuart2015suction}
H.~S. Stuart, M.~Bagheri, S.~Wang, H.~Barnard, A.~L. Sheng, M.~Jenkins, and M.~R. Cutkosky, ``Suction helps in a pinch: Improving underwater manipulation with gentle suction flow,'' in \emph{2015 IEEE/RSJ international conference on intelligent robots and systems (IROS)}.\hskip 1em plus 0.5em minus 0.4em\relax IEEE, 2015, pp. 2279--2284.

\bibitem{granular_gripper}
E.~Brown, N.~Rodenberg, J.~Amend, A.~Mozeika, E.~Steltz, M.~R. Zakin, H.~Lipson, and H.~M. Jaeger, ``Universal robotic gripper based on the jamming of granular material,'' \emph{Proceedings of the National Academy of Sciences}, vol. 107, no.~44, pp. 18\,809--18\,814, 2010.

\bibitem{Granular_foot}
E.~Lathrop, I.~Adibnazari, N.~Gravish, and M.~T. Tolley, ``Shear strengthened granular jamming feet for improved performance over natural terrain,'' in \emph{2020 3rd IEEE International Conference on Soft Robotics (RoboSoft)}, 2020, pp. 388--393.

\bibitem{spiny_hand}
S.~Wang, H.~Jiang, T.~Myung~Huh, D.~Sun, W.~Ruotolo, M.~Miller, W.~R. Roderick, H.~S. Stuart, and M.~R. Cutkosky, ``Spinyhand: Contact load sharing for a human-scale climbing robot,'' \emph{Journal of Mechanisms and Robotics}, vol.~11, no.~3, p. 031009, 2019.

\bibitem{hubrobo_gripper}
K.~Nagaoka, H.~Minote, K.~Maruya, Y.~Shirai, K.~Yoshida, T.~Hakamada, H.~Sawada, and T.~Kubota, ``Passive spine gripper for free-climbing robot in extreme terrain,'' \emph{IEEE Robotics and Automation Letters}, vol.~3, no.~3, pp. 1765--1770, 2018.

\bibitem{risk_aware}
Y.~Shirai, X.~Lin, Y.~Tanaka, A.~Mehta, and D.~Hong, ``Risk-aware motion planning for a limbed robot with stochastic gripping forces using nonlinear programming,'' \emph{IEEE Robotics and Automation Letters}, vol.~5, no.~4, pp. 4994--5001, 2020.

\bibitem{wheel_gripper}
M.~Uda, K.~Sawa, K.~Uno, T.~Kato, L.~T.~L. Zheng, and K.~Yoshida, ``Development and grapsing performance evaluation of a wheel-gripper transformable mechanism,'' in \emph{The Robotics Society of Japan}.\hskip 1em plus 0.5em minus 0.4em\relax RSJ, 2024, pp. 580--583.

\bibitem{GOAT}
Y.~Tanaka, Y.~Shirai, Z.~Lacey, X.~Lin, J.~Liu, and D.~Hong, ``An under-actuated whippletree mechanism gripper based on multi-objective design optimization with auto-tuned weights,'' in \emph{2021 IEEE/RSJ International Conference on Intelligent Robots and Systems}, 2021, pp. 6139--6146.

\bibitem{scaler-b}
Y.~Tanaka, A.~Schperberg, A.~Zhu, and D.~Hong, ``Scaler-b: A multi-modal versatile robot for simultaneous locomotion and grasping,'' in \emph{IEEE International Conference on Robotics and Automation @ 40 (ICRA@40)}.

\bibitem{anymal}
M.~Hutter, C.~Gehring, D.~Jud, A.~Lauber, C.~D. Bellicoso, V.~Tsounis, J.~Hwangbo, K.~Bodie, P.~Fankhauser, M.~Bloesch \emph{et~al.}, ``Anymal-a highly mobile and dynamic quadrupedal robot,'' in \emph{2016 IEEE/RSJ international conference on intelligent robots and systems}.\hskip 1em plus 0.5em minus 0.4em\relax IEEE, 2016, pp. 38--44.

\bibitem{optistate}
A.~Schperberg, Y.~Tanaka, S.~Mowlavi, F.~Xu, B.~Balaji, and D.~Hong, ``Optistate: State estimation of legged robots using gated networks with transformer-based vision and kalman filtering,'' in \emph{2024 IEEE International Conference on Robotics and Automation (ICRA)}, 2024, pp. 6314--6320.

\bibitem{gelslim_iFEM}
D.~Ma, E.~Donlon, S.~Dong, and A.~Rodriguez, ``Dense tactile force estimation using gelslim and inverse fem,'' in \emph{2019 International Conference on Robotics and Automation (ICRA)}, 2019, pp. 5418--5424.

\bibitem{li2024acoustac}
M.~S. Li and H.~S. Stuart, ``Acoustac: Tactile sensing with acoustic resonance for electronics-free soft skin,'' \emph{Soft Robotics}, 2024.

\bibitem{multi_modal_foot_sense}
T.~Tyler, V.~Malhotra, A.~Montague, Z.~Zhao, F.~L. HammondIII, and Y.~Zhao, ``Integrating reconfigurable foot design, multi-modal contact sensing, and terrain classification for bipedal locomotion,'' \emph{IFAC-PapersOnLine}, vol.~56, no.~3, pp. 523--528, 2023.

\bibitem{hall_effect_tactile}
D.~Jones, L.~Wang, A.~Ghanbari, V.~Vardakastani, A.~E. Kedgley, M.~D. Gardiner, T.~L. Vincent, P.~R. Culmer, and A.~Alazmani, ``Design and evaluation of magnetic hall effect tactile sensors for use in sensorized splints,'' \emph{Sensors}, vol.~20, no.~4, p. 1123, 2020.

\bibitem{hall_effect_3d_force}
S.~D.~M. Nasab, A.~Beiranvand, M.~T. Masouleh, F.~Bahrami, and A.~Kalhor, ``Design and development of a multi-axis force sensor based on the hall effect with decouple structure,'' \emph{Mechatronics}, vol.~84, p. 102766, 2022.

\bibitem{magnetic_joystick_3d}
\BIBentryALTinterwordspacing
P.~Malagò, F.~Slanovc, S.~Herzog, S.~Lumetti, T.~Schaden, A.~Pellegrinetti, M.~Moridi, C.~Abert, D.~Suess, and M.~Ortner, ``Magnetic position system design method applied to three-axis joystick motion tracking,'' \emph{Sensors}, vol.~20, no.~23, 2020. [Online]. Available: \url{https://www.mdpi.com/1424-8220/20/23/6873}
\BIBentrySTDinterwordspacing

\bibitem{PolymorphicBlocks}
\BIBentryALTinterwordspacing
R.~Lin, R.~Ramesh, C.~Chi, N.~Jain, R.~Nuqui, P.~Dutta, and B.~Hartmann, ``Polymorphic blocks: Unifying high-level specification and low-level control for circuit board design,'' in \emph{Proceedings of the 33rd Annual ACM Symposium on User Interface Software and Technology}, ser. UIST '20.\hskip 1em plus 0.5em minus 0.4em\relax New York, NY, USA: Association for Computing Machinery, 2020, p. 529–540. [Online]. Available: \url{https://doi.org/10.1145/3379337.3415860}
\BIBentrySTDinterwordspacing

\bibitem{PolymorphicBlocksEdge}
\BIBentryALTinterwordspacing
R.~Lin, R.~Ramesh, N.~Jain, J.~Koe, R.~Nuqui, P.~Dutta, and B.~Hartmann, ``Weaving schematics and code: Interactive visual editing for hardware description languages,'' in \emph{The 34th Annual ACM Symposium on User Interface Software and Technology}, ser. UIST '21.\hskip 1em plus 0.5em minus 0.4em\relax New York, NY, USA: Association for Computing Machinery, 2021, p. 1039–1049. [Online]. Available: \url{https://doi.org/10.1145/3472749.3474804}
\BIBentrySTDinterwordspacing

\bibitem{magpylib}
\BIBentryALTinterwordspacing
M.~Ortner and L.~G. Coliado Bandeira, ``Magpylib: A free python package for magnetic field computation,'' \emph{SoftwareX}, vol.~11, p. 100466, 2020. [Online]. Available: \url{https://www.sciencedirect.com/science/article/pii/S2352711020300170}
\BIBentrySTDinterwordspacing

\bibitem{grbf}
G.~E. Fasshauer and M.~J. McCourt, ``Stable evaluation of gaussian radial basis function interpolants,'' \emph{SIAM Journal on Scientific Computing}, vol.~34, no.~2, pp. A737--A762, 2012.

\bibitem{chung2014empiricalevaluationgatedrecurrent}
J.~Chung, C.~Gulcehre, K.~Cho, and Y.~Bengio, ``Empirical evaluation of gated recurrent neural networks on sequence modeling,'' \emph{arXiv preprint arXiv:1412.3555}, 2014.

\bibitem{dnn_normalize}
L.~Huang, J.~Qin, Y.~Zhou, F.~Zhu, L.~Liu, and L.~Shao, ``Normalization techniques in training dnns: Methodology, analysis and application,'' \emph{IEEE transactions on pattern analysis and machine intelligence}, vol.~45, no.~8, pp. 10\,173--10\,196, 2023.

\end{thebibliography}

\end{document}